\RequirePackage{amsthm}
\documentclass[iicol]{sn-jnl}% Math and Physical Sciences Numbered Reference Style 
\usepackage{graphicx}%
\usepackage{multirow}%
\usepackage{amsmath,amssymb,amsfonts}%
\usepackage{mathrsfs}%
\usepackage[title]{appendix}%
\usepackage{xcolor}%
\usepackage{textcomp}%
\usepackage{manyfoot}%
\usepackage{booktabs}%
\usepackage{algorithm}%
\usepackage{algorithmicx}%
\usepackage{algpseudocode}%
\usepackage{listings}%

\usepackage{booktabs}
\usepackage{multirow}
\usepackage{adjustbox}
\usepackage{graphicx}
\usepackage{wrapfig,lipsum,booktabs}
\usepackage{tabularx}
\usepackage{enumerate} % for roman enumerate in method
\usepackage{type1cm} %for removing type size warnings, mostly

\usepackage{xspace}
\usepackage{xcolor}

\theoremstyle{thmstyleone}%
\theoremstyle{thmstyletwo}%

\theoremstyle{thmstylethree}%

\begin{document}

\title[Article Title]{\quad\quad\quad\quad\quad A Likelihood Ratio-Based Approach \quad\quad\quad\quad\quad \\to Segmenting Unknown Objects}


\author[1,2]{\fnm{Nazir} \sur{Nayal}}\email{nnayal21@ku.edu.tr}
\equalcont{These authors contributed equally to this work.}
\author[3,4]{\fnm{Youssef} \sur{Shoeb}}\email{youssef.shoeb@continental.com}
\equalcont{These authors contributed equally to this work.}

\author*[1,2]{\fnm{Fatma} \sur{Güney}}\email{fguney@ku.edu.tr}

\affil*[1]{\orgdiv{Computer Engineering Department}, \orgname{Koç University}, \country{Turkey}}

\affil[2]{ \orgname{KUIS AI Center}, \orgaddress{\country{Turkey}}}

\affil[3]{\orgname{Continental AG}, \country{Germany}}

\affil[4]{\orgname{Technische Universität Berlin}, \country{Germany}}

\newcommand{\Perp}{\perp\!\!\! \perp}
\newcommand{\bK}{\mathbf{K}}
\newcommand{\bX}{\mathbf{X}}
\newcommand{\bY}{\mathbf{Y}}
\newcommand{\bk}{\mathbf{k}}
\newcommand{\bx}{\mathbf{x}}
\newcommand{\by}{\mathbf{y}}
\newcommand{\bhy}{\hat{\mathbf{y}}}
\newcommand{\bty}{\tilde{\mathbf{y}}}
\newcommand{\bG}{\mathbf{G}}
\newcommand{\bI}{\mathbf{I}}
\newcommand{\bg}{\mathbf{g}}
\newcommand{\bS}{\mathbf{S}}
\newcommand{\bs}{\mathbf{s}}
\newcommand{\bM}{\mathbf{M}}
\newcommand{\bw}{\mathbf{w}}
\newcommand{\eye}{\mathbf{I}}
\newcommand{\bU}{\mathbf{U}}
\newcommand{\bV}{\mathbf{V}}
\newcommand{\bW}{\mathbf{W}}
\newcommand{\bn}{\mathbf{n}}
\newcommand{\bv}{\mathbf{v}}
\newcommand{\bwv}{\mathbf{wv}}
\newcommand{\bq}{\mathbf{q}}
\newcommand{\bR}{\mathbf{R}}
\newcommand{\bi}{\mathbf{i}}
\newcommand{\bj}{\mathbf{j}}
\newcommand{\bp}{\mathbf{p}}
\newcommand{\bt}{\mathbf{t}}
\newcommand{\bJ}{\mathbf{J}}
\newcommand{\bu}{\mathbf{u}}
\newcommand{\bB}{\mathbf{B}}
\newcommand{\bD}{\mathbf{D}}
\newcommand{\bz}{\mathbf{z}}
\newcommand{\bP}{\mathbf{P}}
\newcommand{\bC}{\mathbf{C}}
\newcommand{\bA}{\mathbf{A}}
\newcommand{\bZ}{\mathbf{Z}}
\newcommand{\bff}{\mathbf{f}}
\newcommand{\bF}{\mathbf{F}}
\newcommand{\bo}{\mathbf{o}}
\newcommand{\bO}{\mathbf{O}}
\newcommand{\bc}{\mathbf{c}}
\newcommand{\bm}{\mathbf{m}}
\newcommand{\bT}{\mathbf{T}}
\newcommand{\bQ}{\mathbf{Q}}
\newcommand{\bL}{\mathbf{L}}
\newcommand{\bl}{\mathbf{l}}
\newcommand{\ba}{\mathbf{a}}
\newcommand{\bE}{\mathbf{E}}
\newcommand{\bH}{\mathbf{H}}
\newcommand{\bd}{\mathbf{d}}
\newcommand{\br}{\mathbf{r}}
\newcommand{\be}{\mathbf{e}}
\newcommand{\bb}{\mathbf{b}}
\newcommand{\bh}{\mathbf{h}}
\newcommand{\bhh}{\hat{\mathbf{h}}}
\newcommand{\btheta}{\boldsymbol{\theta}}
\newcommand{\bTheta}{\boldsymbol{\Theta}}
\newcommand{\bpi}{\boldsymbol{\pi}}
\newcommand{\bphi}{\boldsymbol{\phi}}
\newcommand{\bpsi}{\boldsymbol{\psi}}
\newcommand{\bPhi}{\boldsymbol{\Phi}}
\newcommand{\bmu}{\boldsymbol{\mu}}
\newcommand{\bsigma}{\boldsymbol{\sigma}}
\newcommand{\bSigma}{\boldsymbol{\Sigma}}
\newcommand{\bGamma}{\boldsymbol{\Gamma}}
\newcommand{\bbeta}{\boldsymbol{\beta}}
\newcommand{\bomega}{\boldsymbol{\omega}}
\newcommand{\blambda}{\boldsymbol{\lambda}}
\newcommand{\bLambda}{\boldsymbol{\Lambda}}
\newcommand{\bkappa}{\boldsymbol{\kappa}}
\newcommand{\btau}{\boldsymbol{\tau}}
\newcommand{\balpha}{\boldsymbol{\alpha}}
\newcommand{\nR}{\mathbb{R}}
\newcommand{\nN}{\mathbb{N}}
\newcommand{\nL}{\mathbb{L}}
\newcommand{\nE}{\mathbb{E}}
\newcommand{\cN}{\mathcal{N}}
\newcommand{\cM}{\mathcal{M}}
\newcommand{\cR}{\mathcal{R}}
\newcommand{\cB}{\mathcal{B}}
\newcommand{\cL}{\mathcal{L}}
\newcommand{\cH}{\mathcal{H}}
\newcommand{\cS}{\mathcal{S}}
\newcommand{\cT}{\mathcal{T}}
\newcommand{\cO}{\mathcal{O}}
\newcommand{\cC}{\mathcal{C}}
\newcommand{\cP}{\mathcal{P}}
\newcommand{\cE}{\mathcal{E}}
\newcommand{\cI}{\mathcal{I}}
\newcommand{\cF}{\mathcal{F}}
\newcommand{\cK}{\mathcal{K}}
\newcommand{\cY}{\mathcal{Y}}
\newcommand{\cX}{\mathcal{X}}
\def\bgamma{\boldsymbol\gamma}

\newcommand{\specialcell}[2][c]{%
  \begin{tabular}[#1]{@{}c@{}}#2\end{tabular}}

\newcommand{\figref}[1]{\Fig~\ref{#1}}
\newcommand{\secref}[1]{Section~\ref{#1}}
\newcommand{\eqnref}[1]{Eq.~\ref{#1}}
\newcommand{\tabref}[1]{Table~\ref{#1}}

\newcommand{\rulesep}{\unskip\ \vrule\ }

%\DeclareMathOperator*{\argmax}{argmax~}
% \DeclareMathOperator*{\argmin}{argmin~}

% KL divergence
%\DeclarePairedDelimiterX{\infdivx}[2]{[}{]}{%
%  #1\;\delimsize\|\;#2%
%}
%\newcommand{\infdiv}{D\infdivx}

% Kullback-Leibler divergence (or relative entropy)
\newcommand{\KLD}[2]{D_{\mathrm{KL}} \Big(#1 \mid\mid #2 \Big)}

\renewcommand{\b}{\ensuremath{\mathbf}}

\def\mc{\mathcal}
\def\mb{\mathbf}

\newcommand{\T}{^{\raisemath{-1pt}{\mathsf{T}}}}

\makeatletter
\DeclareRobustCommand\onedot{\futurelet\@let@token\@onedot}
\def\@onedot{\ifx\@let@token.\else.\null\fi\xspace}
\def\eg{e.g\onedot} \def\Eg{E.g\onedot}
\def\ie{i.e\onedot} \def\Ie{I.e\onedot}
\def\cf{cf\onedot} \def\Cf{Cf\onedot}
\def\etc{etc\onedot} \def\vs{vs\onedot}
\def\wrt{wrt\onedot}
\def\dof{d.o.f\onedot}
\def\etal{et~al\onedot} \def\iid{i.i.d\onedot}
\def\Fig{Fig\onedot} \def\Eqn{Eqn\onedot} \def\Sec{Sec\onedot} \def\Alg{Alg\onedot}
\makeatother

\newcommand{\xdownarrow}[1]{%
  {\left\downarrow\vbox to #1{}\right.\kern-\nulldelimiterspace}
}

\newcommand{\xuparrow}[1]{%
  {\left\uparrow\vbox to #1{}\right.\kern-\nulldelimiterspace}
}

\definecolor{mygreen}{RGB}{93,173,85}
\newcommand{\improv}[2]{
{#1} \fontsize{7.5pt}{1em}\selectfont\color{mygreen}{$\!\uparrow\!$ \textbf{#2}}
}

\newcommand{\decrease}[2]{
{#1} \fontsize{7.5pt}{1em}\selectfont\textcolor{red}{$\!\downarrow\!$ \textbf{#2}}
}

% nice url font and color
% \renewcommand\UrlFont{\color{blue}\rmfamily}

\newcommand\todo[1]{\textcolor{red}{#1}}

% rotation
\newcommand*\rot{\rotatebox{90}}
\newcommand{\boldparagraph}[1]{\vspace{0.15cm}\noindent{\bf #1:} }
\newcommand{\boldquestion}[1]{\vspace{0.15cm}\noindent{\bf #1}? }

\newcommand{\ftm}[1]{ \noindent {\color{magenta} {#1}} }
\newcommand{\nn}[1]{ \noindent {\color{blue} {#1}} }
\newcommand{\my}[1]{ \noindent {\color{orange} {\bf Youssef:} {#1}} }

\newcommand{\red}[1]{{\color{red}#1}}

\abstract{}
\abstract{
Addressing the Out-of-Distribution (OoD) segmentation task is a prerequisite for perception systems operating in an open-world environment. Large foundational models are frequently used in downstream tasks, however, their potential for OoD remains mostly unexplored. We seek to leverage a large foundational model to achieve robust representation. Outlier supervision is a widely used strategy for improving OoD detection of the existing segmentation networks. However, current approaches for outlier supervision involve retraining parts of the original network, which is typically disruptive to the model's learned feature representation. Furthermore, retraining becomes infeasible in the case of large foundational models. Our goal is to retrain for outlier segmentation without compromising the strong representation space of the foundational model. 
To this end, we propose an adaptive, lightweight unknown estimation module (UEM) for outlier supervision that significantly enhances the OoD segmentation performance without affecting the learned feature representation of the original network. UEM learns a distribution for outliers and a generic distribution for known classes. Using the learned distributions, we propose a likelihood-ratio-based outlier scoring function that fuses the confidence of UEM with that of the pixel-wise segmentation inlier network to detect unknown objects. We also propose an objective to optimize this score directly. Our approach achieves a new state-of-the-art across multiple datasets, outperforming the previous best method by 5.74\% average precision points while having a lower false-positive rate. Importantly, strong inlier performance remains unaffected.
}




\maketitle

%-------------------------------------------------------------------------
\section{Introduction}
\label{sec:intro}
Semantic segmentation represents a significant advancement in deep learning. Learned features are densely mapped to a pre-defined set of classes by a pixel-level classifier. The remarkable performance of end-to-end models on this closed set has led researchers to consider the next challenge: extending semantic segmentation to the open-world setting where objects of unknown classes also need to be segmented. One of the biggest challenges in segmenting unknown objects is the lack of outlier data.

In this work, we first attack the lack of data for unknown segmentation by utilizing a large foundation model, DINOv2~\cite{Oquab2024TMLR}, for a robust representation space. The availability of internet-scale data has enabled the training of large visual foundation models, known for their generalization capabilities across various tasks~\cite{Zhang2023NeurIPS, Blumenkamp2024CORL, Aydemir2023NeurIPS, Nguyen2023CVPR}. 
Despite these promising generalization capabilities, their potential for unknown object segmentation remains mostly unexplored. Only recently, PixOOD~\cite{Vojir2024ECCV} has used DINOv2 without any training to avoid biases in industrial settings, however, their performance falls significantly behind the methods that use outlier supervision on commonly used SMIYC benchmark.

While collecting representative data for all possible classes in an open-world setting is impracticable, existing methods perform significantly better when trained using proxy outlier data~\cite{Grcic23CVPRW, Nayal2023ICCV, Rai2023ICCV}, for example, obtained with the cut-and-paste method. Retraining with outlier supervision improves unknown segmentation but causes problems for known classes due to the reshaping of the representation space. Furthermore, retraining the entire model becomes infeasible in the case of large foundational models.
We propose a novel way of utilizing proxy outlier data to improve the segmentation of unknown classes without compromising the performance of known classes.

Semantic segmentation models are typically trained to predict class probabilities with a softmax classifier. With a cross-entropy loss on the predicted class probabilities, the model learns to discriminate features of a certain class from the others. Such models excel in learning \emph{discriminative} representations for the known classes but struggle to generalize to unknown classes due to partitioning the feature space between known classes. As an alternative, deep generative models directly learn a density model to predict the likelihood of a data sample. This likelihood is expected to be lower for outliers, such as samples from unknown classes. However, generative models often require more computational resources and can be challenging to train effectively.

Due to their potential to learn well-calibrated scores, deep generative models have been widely explored for out-of-distribution (OoD) tasks. However, in segmentation, their performance is often inferior to that of discriminative counterparts \cite{Lee2018NeurIPS, Haldimann2019ARXIV, Xia2020ECCV, Vojir2021ICCV}. To benefit from the best of both worlds, GMMSeg~\cite{Liang2022NeurIPS} presents a hybrid approach by augmenting the GMM-based generative model with discriminatively learned features. While discriminative features boost the inlier performance, GMM helps achieve an impressive OoD performance without explicitly training for it.

Nalisnick \etal~\cite{Nalisnick2019ICLR} test the ability of deep generative models to detect OoD. They show that a generative model trained on one dataset assigns higher likelihoods to samples from another than those from the training dataset itself. Zhang \etal~\cite {Zhang2022NeurIPSW} first explain this phenomenon by showing that the expected log-likelihood is mathematically larger for out-of-distribution data and then propose to differentiate between outlier and OoD detection. While the learned density function can be used to detect outliers with respect to a single distribution, \emph{OoD detection requires comparing two distributions}. As initially proposed by Bishop~\cite{Bishop1994}, OoD detection can be considered model selection between in-distribution and out-of-distribution data. Although an out-of-distribution is not often explicitly modeled, Zhang \etal~\cite {Zhang2022NeurIPSW} show that several existing works in OoD perform a likelihood ratio test with a proxy distribution for OoD, \eg from auxiliary OoD datasets~\cite{Hendrycks2019ICLR} or using background statistics~\cite{Ren2019NeurIPS}.

In this paper, we propose applying the likelihood ratio as a principled way of detecting OoD in semantic segmentation. To calculate the likelihood ratio, we propose to train a lightweight unknown estimation module (UEM) on top of an already trained semantic segmentation model with a fixed number of semantic classes. UEM estimates an OoD distribution using proxy outlier data and a class-agnostic inlier distribution to calculate the likelihood ratio score.
We also propose an objective to optimize the likelihood ratio score and train UEM with this objective. We show that our formulation is general enough to apply to both discriminative and generative segmentation models, with an example for each in the experiments. Our proposed method achieves state-of-the-art performance on multiple benchmarks while maintaining the same inlier performance.

\section{Related Work}
\label{sec:related}

\boldparagraph{OoD without Outlier Data}
Earlier approaches for OoD detection rely on uncertainty estimation methods to model predictive uncertainty. The uncertainty of a model can be estimated through maximum softmax probabilities~\cite{Hendrycks2017ICLR}, ensembles~\cite{Lakshminarayanan2017NeurIPS}, MC-dropout \cite{Gal2016ICML}, or by learning to estimate the confidence directly \cite{Kendall2017NeurIPS}. However, posterior probabilities in a closed-set setting may not always be well-calibrated for an open-world setting, potentially leading to overly confident predictions for unfamiliar categories \cite{Guo2017ICML, Jiang2018NeurIPS, Minderer2021NeurIPS}.

\boldparagraph{OoD with Outlier Data}
Hendrycks \etal~\cite{Hendrycks2019ICLR} introduce outlier exposure to improve OoD detection. Outlier exposure leverages a proxy dataset of outliers to discover signals and learn heuristics for OoD samples. Chan \etal~\cite{Chan2021ICCV} use a proxy dataset and entropy maximization to fine-tune the model to give high entropy scores to unknown samples. Similarly, RbA~\cite{Nayal2023ICCV} uses a proxy dataset to fine-tune the model to produce low logit scores on unknown objects. We follow a similar approach in our work and use a proxy dataset to learn a proxy distribution of OoD. However, our proxy dataset is only used to adjust the parameters of a small discriminator model, so it does not affect the performance of the inlier model.

\boldparagraph{Deep Generative Models for OoD}
Generative models have been used to identify outliers based on the estimated probability density of the inlier training data distribution. Liang \etal~\cite{Liang2022NeurIPS} use a mixture of Gaussians to represent the data distribution within each class and model OoD instances as low-density regions. Other methods use normalizing flow~\cite{Blum2021IJCV, Grcic2024SENSORS} or an energy-based model~\cite{Grcic2022ECCV} to estimate inlier data density. However, estimating a data density of inliers only does not behave as expected for OoD detection, as Nalisnick \etal~\cite{Nalisnick2019ICLR} show in their analysis of several deep generative models.
Instead of a single density estimation, we treat OoD detection as model selection between two distributions as proposed in \cite{Zhang2022NeurIPSW}. We directly train the model to optimize the likelihood ratio between an in-distribution and an out-of-distribution for a better separation of outliers. To our knowledge, this is the first work to consider the likelihood ratio for segmenting outliers.    

\boldparagraph{Mask-Based OoD} A recent trend in OoD segmentation is to use mask-based models by predicting and classifying masks~\cite{Cheng2021NeurIPS, Cheng2022CVPR, Li2023CVPR}. 
In masked-based models such as Mask2Former~\cite{Cheng2022CVPR}, each query specializes in detecting a certain known class~\cite{Nayal2023ICCV, Ackermann2023BMVC}. Based on this property of mask-based models, RbA~\cite{Nayal2023ICCV} proposes an outlier scoring function based on the probability of not belonging to any known classes. Utilizing the same property, Maskomaly~\cite{Ackermann2023BMVC} selects outlier masks by thresholding the per-class mIoU on a validation set. Mask2Anomaly~\cite{Rai2023ICCV} augments Mask2Former with a global masked-attention mechanism and trains it using a contrastive loss on outlier data.
EAM~\cite{Grcic23CVPRW} performs OoD detection via an ensemble over mask-level scores.
Almost all of these methods, except for Maskomaly~\cite{Ackermann2023BMVC}, which is a simple inference-time post-processing technique, show the importance of utilizing OoD data during training. In this paper, we propose a better way of utilizing outlier data with the likelihood ratio, outperforming mask-based models in most metrics with pixel-based classification.

\boldparagraph{Foundational Models for OoD} Foundational models trained on large datasets have shown impressive zero-shot performance on downstream tasks like classification and segmentation \cite{Radford2021ICML, Oquab2024TMLR, Ranzinger2024CVPR}. For image-level OoD classification, Vojir \etal~\cite{Vojir2023ICCVW} leverage generic pre-trained representation from CLIP~\cite{Radford2021ICML}. Wang \etal~\cite{Wang2023ICCV} train a negation text-encoder to equip CLIP with the ability to separate OoD samples from in-distribution samples. Recently, PixOoD~\cite{Vojir2024ECCV} utilizes DINOv2~\cite{ Oquab2024TMLR} for modeling the in-distribution data and achieves competitive results for OoD segmentation without using any outlier training. Initial work started exploring the potential of foundational models for OoD by building on their powerful representations. In this work, we take it further and improve outlier performance by retraining with outlier supervision without affecting the representation space of the foundational model.

\begin{figure*}[tb]
  \centering
  \includegraphics[width=\linewidth]{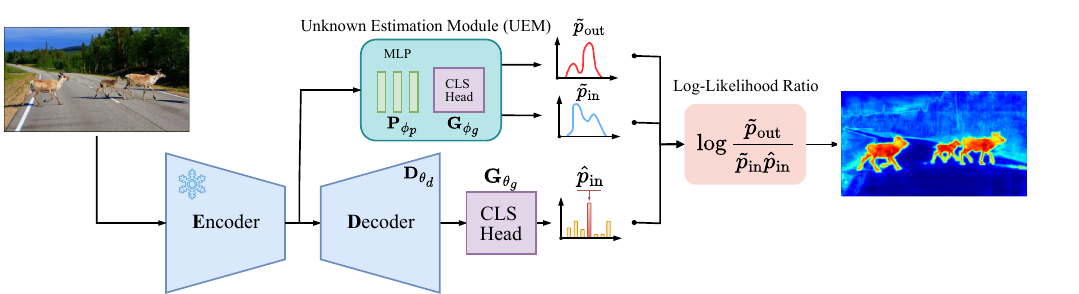}
  \caption{\textbf{Overview.} Our proposed unknown estimation module (UEM) takes the input from the frozen encoder backbone and learns the outlier and inlier distributions $\tilde{p}_{\text{out}}$ and $\tilde{p}_{\text{in}}$. Then, we calculate the log-likelihood ratio by combining the outputs of UEM with the class probabilities of the inlier model $\hat{p}_{\text{in}}$.
  }
  \label{fig:overview}
\end{figure*}
\section{Methodology}
\label{sec:method}

\subsection{Overview}
We propose a two-stage approach: In the first stage, a semantic segmentation model is trained solely on the known data with the standard segmentation losses. In the second stage, the semantic segmentation model is fully frozen to maintain its exact inlier performance. We train an adaptive, lightweight unknown estimation module that estimates $\tilde{p}_{\text{out}}$ and a generic inlier distribution $\tilde{p}_{\text{in}}$ after injecting the training dataset with pseudo-unknown pixels in the second stage. With this setup, we propose an OoD scoring function based on the likelihood ratio by combining the output of this module and the inlier part and propose a loss function to optimize it. In \figref{fig:overview}, we provide an overview of our proposed approach.

\subsection{Notation and Preliminaries}
Given an input image $\bx \in \nR^{3 \times H \times W}$ and its corresponding label map $\by \in \cY^{H \times W}$, a closed-set semantic segmentation model learns a mapping from the input pixels to the class logits $\bF_{\theta}(\bx): \mathbb{R}^{3 \times H \times W} \rightarrow \mathbb{R}^{K \times H \times W}$, where $\cY = \{1, \dots, K\}$ is the set of known class labels during training. 
In OoD segmentation, we extend the label space to $\cY^{'} = \cY \cup \{K + 1\}$, where $K + 1$ represents semantic categories unseen during training or the OoD class. To identify pixels belonging to the class $K + 1$, we define a scoring function $\cS_{\text{out}}(\bx) \in \mathbb{R}^{H \times W}$ that assigns high values to OoD pixels and low values to inlier pixels belonging to $\cY$.

\boldparagraph{Likelihood Ratio} Previous work in image-level OoD detection~\cite{Nalisnick2019ICLR} has shown that when $\cS_{\text{out}}(x)$ for an image is defined using the likelihood density of the training data, it assigns high likelihood values to some OoD samples. This limitation of likelihood-based methods has been mitigated by defining $\cS_{\text{out}}(x)$ as the likelihood ratio (LR) between two distributions: $p_{\text{in}}$ representing the likelihood of the sample belonging to the inlier distribution, and $p_{\text{out}}$ representing the likelihood of a pixel $x$ is an outlier. Formally:
\begin{equation}
    \text{LR}(x) ~=~ \frac{p_{\text{out}}(x)}{p_{\text{in}}(x)}
    \label{eq:lr}
\end{equation}
In this formulation, the likelihood of a sample being an inlier is reinforced by the likelihood of it not being an outlier, and vice versa. While defining $p_{\text{in}}$ is done using the inlier dataset, defining $p_{\text{out}}$ is challenging due to the unbounded diversity of $p_{\text{out}}$ compared to $p_{\text{in}}$. Therefore, using different assumptions, previous work explores approximating $p_{\text{out}}$~\cite{Ren2019NeurIPS, Zhang2021NeurIPS}. In this work, we explore representing $p_{\text{out}}$ by means of both a generative and discriminative class-conditional distribution. We achieve these by utilizing synthetic pseudo-OoD objects based on driving scenes as done in previous works~\cite{Rai2023ICCV, Nayal2023ICCV, Grcic23CVPRW}.

\subsection{Learning an Inlier Segmentor}
\label{sec:inlier_segmentor}

The existing pixel-level inlier segmentation models typically consist of three parts: 

\begin{enumerate}[i.]
    \item a feature extractor $\bE: \nR^{3 \times H \times W} \mapsto \mathbb{R}^{C_e \times \dot{H} \times \dot{W}}$, reducing spatial dimension to $\dot{H} \times \dot{W}$,
    \item a decoder $\bD_{\theta_d}: \nR^{C_e \times \dot{H} \times \dot{W}} \mapsto \nR^{C_d \times H \times W}$, increasing it back to the original $H \times W$, and
    \item a classification head $\bG_{\theta_g} : \nR^{C_d \times H \times W} \mapsto \mathbb{R}^{K \times H \times W}$ mapping features to class logit scores.
\end{enumerate} 
$C_e$ and $C_d$ denote the encoder and decoder's hidden dimension size, respectively. Hence, the mapping $\bF_{\theta}(\bx): \nR^{3 \times H \times W} \mapsto \nR^{K \times H \times W}$ is defined as $\bF_{\theta} = \bE \circ \bD_{\theta_d} \circ \bG_{\theta_g}$. In this notation, $\theta$ is the set of all learnable parameters and contains the union of $\theta_d$ and $\theta_g$. In some cases, features from multiple layers of the encoder $\bE$ can be passed on to the decoder $\bD$ to process features in a multi-scale fashion. We omit this in the notation for simplicity. 

For the backbone, we use DINOv2~\cite{Oquab2024TMLR}, which is a self-supervised ViT \cite{dosovitskiy2021an} that has been shown to produce robust and rich visual representations~\cite{Ranzinger2024CVPR}. To maintain its rich representation, we freeze the backbone throughout all stages of training. 
For the decoder, we utilize a standard Feature Pyramid Network (FPN)~\cite{Lin2017CVPR} that takes features from multiple layers of the encoder and fuses them to produce an output feature map. 
For the classification head, we explore using two types of classifiers: generative and discriminative. Although the discriminative version seems less suitable for likelihood computations, we show that it performs exceptionally well under certain assumptions.

\boldparagraph{Generative Classifier} We adopt the generative classification formula proposed in \cite{Liang2022NeurIPS}, which replaces the linear softmax classification head by learning class densities of each pixel $p(x|k)$ with Gaussian Mixture Models (GMMs), where each class is represented with a separate GMM with a uniform prior on the component weights. Formally:
\begin{equation}
    p(x | k, \theta_g) = \sum_{c=1}^{C}{\pi_{kc}~\cN(x ; \mu_{kc}, \Sigma_{kc})}
\end{equation}
where $C$ is the number of components per GMM, $\pi_{kc}$ is the component mixture weight for component $c$ of class $k$, $\mu_{kc}$,$\Sigma_{kc}$ are the mean and covariance matrix respectively, and $\cN$ is the Gaussian distribution. The GMM parameters are learned with a variant of the Expectation-Maximization (EM) algorithm called Sinkhorn EM, which adds constraints that enforce an even assignment of features to mixture components, thereby improving the training stability. 
For more details, please refer to \cite{Liang2022NeurIPS}.

\boldparagraph{Discriminative Classifier} We train a single linear layer as a discriminative classifier. In this version, the parameters of the model $\theta$ are supervised by the cross-entropy loss:
\begin{equation}
    \theta^{*} = \text{argmin}_{\theta} -\sum_{(x, k) \in \mathcal{D}} \log p(k | x, \theta)
    \label{eq:cross_entropy}
\end{equation}
where $\mathcal{D}$ is the set of image-label pairs, and $p(k | x, \theta)$ is the softmax output of class $k$ after mapping it to class logits first, $\hat{y} = \bF_{\theta}(x)$:
\begin{equation}
    p(k | x, \theta) = \frac{\exp(\hat{y}_k)}{\sum_{k'} \exp(\hat{y}_{k'})}
    \label{eq:prob}
\end{equation}

\subsection{Unknown Estimation Module (UEM)}
At this stage, we assume the existence of an inlier segmentation model trained as described in \secref{sec:inlier_segmentor}. The unknown estimation module (UEM) consists of a projection module $\bP_{\phi_p} \in \nR^{C_p \times H \times W}$, where $C_p$ is the hidden size output for the projection module, which is a 3-layer Multi-Layer Perceptron (MLP) that takes the output of the frozen backbone and produces a projected feature map as follows:
\begin{equation}
    \bP_{\phi_p}(\bx) = \text{MLP}\big(\bE(\bx)\big)
\end{equation}

After that, the projected features are fed to a classification head $\bG_{\phi_g}$ with two classes: one class maps to the OoD distribution and another to a generic inlier distribution learned directly from the backbone. The classifier head can be defined in a generative or discriminative fashion as explained in \secref{sec:inlier_segmentor}. Hence, the output of UEM: $\bU \in \mathbb{R}^{2 \times H \times W}$ is defined as follows:
\begin{equation}
    \bU_{\phi}(\bx) = \bG_{\phi_g}\big(\bP_{\phi_p}(\bx)\big)
\end{equation}

From the outputs of this module, we denote $\tilde{p}_{\text{out}} = \bU_1(\bx)$ and $\tilde{p}_{\text{in}} = \bU_0(\bx)$ as the likelihood of a sample $\bx$ being an outlier or an inlier respectively. Hence, in the case of the generative classifier, the likelihoods would be those of the learned GMMs. In the case of a discriminative classifier, these would correspond to the class posterior probabilities $p(c|x)$ as the output of $\bG_{\phi_g}$. 

\subsection{Log-Likelihood Ratio Score}
First, we outline the formulation assuming a generative classifier for both the inlier model and the UEM module. 

\boldparagraph{Generative} We propose the log-likelihood ratio as an OoD scoring function $\cS_{\text{out}}$ where the likelihood ratio is defined in \eqref{eq:lr}. For this, we need to define $p_{\text{out}}$ and $p_{\text{in}}$. We simply set the outlier distribution $p_{\text{out}} = \tilde{p}_{\text{out}}$,  where $\tilde{p}_{\text{out}}(\bx) \sim \text{GMM}$ with a uniform prior $\pi_c = \frac{1}{C}$:
\begin{equation}
    \begin{split}
        \tilde{p}_{\text{out}}(\bx) &= \sum_{c=1}^{C}\pi_{c} \cN(\bx ;\mu_c, \Sigma_c) \\
        \label{eq:tilde_p_out}
    \end{split}
\end{equation}
where $C$ is the number of components and $\cN$ is the Gaussian distribution. As for the inlier distribution $p_{\text{in}}$, we define it by combining $\tilde{p}_{in}$ with the likelihood that a sample is inlier based on the inlier segmentation model. Due to the independence of the two sources of inlier confidence, $p_{\text{in}}$ can be defined as their product: $p_{\text{in}} = \tilde{p}_{\text{in}}\cdot\hat{p}_{\text{in}}$.
In this case, $\tilde{p}_{\text{in}}(\bx)$ follows the same form of $\tilde{p}_{\text{out}}(\bx)$ in \eqref{eq:tilde_p_out}. As for $\hat{p}_{\text{in}}(\bx)$, we have:
\begin{equation}
    \begin{split}
        \hat{p}_{\text{in}}(\bx) &= \max_k{\log p(k|\bx)} \\
        \label{eq:hat_p_in}
    \end{split}
\end{equation}

However, since the inlier segmentation model is a generative classifier, we have $p(k|\bx) \sim \text{GMM}$. And following \cite{Liang2022NeurIPS}, the log probability is used as class logit scores, hence $\bF_k(\bx) = \log p(k|\bx)$, which makes the full log-likelihood ratio score as follows: 
    \begin{equation}
        \begin{split}
        \text{LLR}(\bx) &=
        \log{\frac{p_{\text{out}}(\bx)}{p_{\text{in}}(\bx)}} \\
        &=\log p_{\text{out}}(\bx) - \log{p_{\text{in}}(\bx)} \\
        &= \log \tilde{p}_{\text{out}}(\bx) - \log{\tilde{p}_{\text{in}}(\bx)} - \max_k \log p(k|\bx) \\
        &= \log \tilde{p}_{\text{out}}(\bx) - \log{\tilde{p}_{\text{in}}(\bx)} - \max_k \bF_k(\bx)
        \end{split}
        \label{eq:llr_g_g}
    \end{equation}

LLR unifies the confidence values of the inlier model and our proposed UEM in a single objective.

\boldparagraph{Discriminative} In this case,  $p_{\text{in}}$ and $p_{\text{out}}$ are defined as $\tilde{p}_{\text{in}}$ and $\tilde{p}_{\text{out}}$ respectively. The difference compared to the generative case is that these terms are not defined as GMMs, but rather as logits computed through a linear classifier. As for $\hat{p}_{\text{in}}$, we define it to be the maximum of class logits defined in \eqref{eq:prob}. Hence,  the LLR score in this case can be written as:

\begin{align}
   \begin{split}
       \text{LLR}(\bx) &= \log{\frac{p_{\text{out}}(\bx)}{p_{\text{in}}(\bx)}} \\
       &= \log p_{\text{out}}(\bx) - \log{p_{\text{in}}(\bx)} \\
       &= \log \tilde{p}_{\text{out}}(\bx) - \log \tilde{p}_{\text{in}}(\bx) - \max_k{\log p(k|x)} \\
       &= \log \tilde{p}_{\text{out}}(\bx)  - \log \tilde{p}_{\text{in}}(\bx)\\
       &\quad- \max_k \Big({\bF_k(\bx)} + \log{\sum_{k' \in \cY}{\exp\big(\bF_{k'}(\bx)\big)}}\Big)
   \end{split}
\end{align}

As the normalization term $ \log{\sum_{k \in \cY}{\exp(\bF(\bx))}}$ does not affect the maximum, we obtain the final form of the scoring function as follows:

\begin{equation}
      \text{LLR}(\bx) = \log \tilde{p}_{\text{out}}(\bx) - \log \tilde{p}_{\text{in}}(\bx) - \max_k{\bF_k(\bx)}
\end{equation}
This shows that the generative and discriminative case formulations converge to the same equation.

\subsection{Log-Likelihood Ratio Loss}
The proposed unknown estimation module $\bU_{\phi}$ is supervised by the LLR loss defined as follows:
\begin{equation}
    \cL_{\text{LLR}}(\bx, \tilde{\by}) = \text{BCE}(\text{LLR}(\bx), \tilde{\by}) + \alpha~\cL_{\text{GMM}}
\end{equation}
where $\tilde{\by}$ is a binary label map denoting known and pseudo-outlier pixels, BCE is the Binary Cross Entropy Loss, and $\cL_{\text{GMM}}$ is the loss used to train the GMM component in case the generative classifier used as in \cite{Liang2022NeurIPS}. 
The $\cL_{\text{GMM}}$ consists of two terms as follows: 
    \begin{equation}
        \cL_{\text{GMM}} = \cL_{\text{CE}} + \beta \cL_{\text{contrast}}
    \end{equation}
where $\beta$ is a weighting coefficient. $\cL_{\text{CE}}$ is the cross-entropy loss which is applied on the output logit scores of the generative classifier head $\bF(\bx)$, which as shown in \cite{Liang2022NeurIPS} corresponds to the class log probabilities of the class GMMs. As for $\cL_{\text{contrast}}$, this loss is applied to contrast between every component within every class GMM with all other components, including those with the same class and of the other classes. In other words, using the Sinkhorn algorithm, each feature in the image is assigned to a unique component within the GMM of its ground truth class. If there are $K$ classes and $C$ components with each $GMM$, then the cross entropy loss optimizes the likelihood feature belonging to its assigned component and minimizes it for the other $CK - 1$ components within the classification head.

\section{Experiments}
\label{sec:experiments}
\subsection{Experimental Setup}

In our experiments, we use an inlier segmentation network composed of a feature extractor, an FPN~\cite{Lin2017CVPR} pixel decoder, and a generative classification head (GMMSeg~\cite{Liang2022NeurIPS}). The feature extractor is frozen, and we train the pixel decoder and segmentation head in the first stage on random patches of size $518 \times 1036$ taken from the Cityscapes dataset~\cite{Cordts2016CVPR}. In the second stage, we train our unknown estimation module using a modified version of Anomaly Mix~\cite{Tian2022ECCV}, where we randomly cut and past objects from the COCO dataset~\cite{Lin2014ECCV} on the training data for outlier supervision. During outlier supervision, all the trained parameters of the main segmentation network are frozen to maintain inlier performance. 
Finally, we maintain the training resolution during inference but with a sliding window approach to cover the whole image. 

\begin{table*}[ht]
    \centering
    \adjustbox{max width=\textwidth}{%
    \begin{tabular}{l l c c c c c c } 
     \toprule
     \multirow{2}{*}{Backbone} & \multirow{2}{*}{mIoU$\uparrow$}  & \multicolumn{3}{c}{Road Anomaly} & \multicolumn{3}{c}{FS LaF}  \\
         \cmidrule(r){3-5} \cmidrule(l){6-8}
        &  & $AUROC$ $\uparrow$ & $AP$ $\uparrow$ & $FPR$ $\downarrow$ & $AUROC$ $\uparrow$ & $AP$ $\uparrow$ & $FPR$ $\downarrow$ \\
         \midrule
        \multirow{1}{*}{Swin-b \cite{Liu2021ICCV}}&
           
        81.6  & 92.80 & 65.42  & 27.96 &  96.65 & 43.53 & 18.88   \\
        %\midrule
        \multirow{1}{*}{CLIP~\cite{Radford2021ICML}} &
         77.8  & 97.84 & 87.84 & 10.81 & 99.95 &  55.84 & 4.12 \\
         %\midrule
        \multirow{1}{*}{DINOv2-b \cite{Oquab2024TMLR}} &
         82.8  & 98.37 & 92.86 & 8.95 & 98.64 &  64.02 & 1.62 \\
         \bottomrule   
        \end{tabular}}
        \caption{\textbf{Ablation of Backbone Feature Extractor.} We compare the inlier and OoD detection performance using different backbones. We find the DINOv2 backbone to offer the best inlier and outlier performance.} 
        \label{tab:inlier_segmentation}
\end{table*}

\boldparagraph{Evaluation Datasets and Metrics}
We report the performance on SMIYC~\cite{Chan2021NeurIPS} Anomaly Track (SMIYC-AT), Obstacle Track (SMIYC-OT), RoadAnomaly~\cite{Lis2019ICCV}, and the validation set of Fishyscapes LostandFound (FS LaF)~\cite{Blum2021IJCV}. SMIYC-AT and RoadAnomaly are real-world images featuring one or several OoD objects of varying sizes and categories. SMIYC-OT and FS LaF assess the model's capability to identify small-sized obstacles on the road. 
We evaluate the performance of our method using common pixel-wise anomaly segmentation metrics: Average Precision ($AP$) and False Positive Rate ($FPR$) at True Positive Rate of 95\%.

\subsection{Quantitative Results}
\boldparagraph{Backbone Feature Extractor} The proposed Unknown Estimation Module (UEM) builds on a strong backbone model as the feature extractor. 
The backbone plays a critical role by encoding images into a rich representation space, which helps first model the inliers and then differentiate the outliers with the UEM.
We compare the performance of three different backbones for feature extraction, including a self-supervised one, DINOv2~\cite{Oquab2024TMLR}; a contrastive one, CLIP~\cite{Radford2021ICML}; and a supervised one, Hierarchical Swin Transformer~\cite{Liu2021ICCV} for the baseline segmentation network.

Table \ref{tab:inlier_segmentation} shows the mIoU performance of the inlier network using different backbones on Cityscapes and the anomaly segmentation performance of UEM trained on top of the inlier network on RoadAnomaly and Fishyscapes. While DINO and Swin show comparable performance on inlier data, DINO significantly outperforms Swin in handling outliers. CLIP shows lower inlier performance than both but surpasses Swin in outlier detection. This difference in outlier performance can be attributed to the pre-training of DINO and CLIP on larger and more diverse datasets, which results in more robust feature representations capable of effectively modeling both in-distribution and out-of-distribution.

\begin{table*}[ht]
    \centering
    \adjustbox{max width=\textwidth}{%
    \begin{tabular}{l l c c c c c c } 
     \toprule
     \multirow{2}{*}{Backbone} & \multirow{2}{*}{Scoring}  & \multicolumn{3}{c}{Road Anomaly} & \multicolumn{3}{c}{FS LaF}  \\
        \cmidrule(r){3-5} \cmidrule(l){6-8}
        &  & $AUROC$ $\uparrow$ & $AP$ $\uparrow$ & $FPR$ $\downarrow$ & $AUROC$ $\uparrow$ & $AP$ $\uparrow$ & $FPR$ $\downarrow$ \\
         \midrule
        \multirow{3}{*}{Swin-b \cite{Liu2021ICCV}}&
         ID  & 80.14 & 32.34 &56.53 &82.54 &5.51 &69.16   	          \\
         & OoD  & \improv{91.63}{11.49} &\improv{61.48}{29.14} 	& \improv{32.34}{24.19}  & \improv{96.46}{13.92}  & \improv{38.30}{32.78} & \improv{18.10}{51.06} \\  
         & LR  & \improv{92.80}{12.66} & \improv{65.42}{33.08}  & \improv{27.96}{28.57} & \improv{96.65}{14.11} & \improv{43.53}{38.02} & \improv{18.88}{50.28}   \\
        \midrule
        \multirow{3}{*}{CLIP~\cite{Radford2021ICML}} &
        ID  & 90.35 & 50.71 & 34.72 & 91.97 & 13.28 & 38.70\\
        & OoD  & \improv{96.78}{6.43} &  \improv{83.54}{32.83}&  \improv{17.41}{17.31} & \improv{98.56}{6.59} &  \improv{35.70}{22.42} & \improv{6.05}{32.65} \\
         & LR  & \improv{97.84}{7.49} & \improv{87.84}{37.13} & \improv{10.81}{23.91} & \improv{99.95}{7.98} &  \improv{55.84}{42.56} & \improv{4.12}{34.58} \\
         \midrule
        \multirow{3}{*}{DINOv2-b \cite{Oquab2024TMLR}} &
        ID  & 92.94 & 64.77 & 28.25 & 90.83 & 15.00 & 40.70\\
        & OoD  & \improv{97.45}{4.51} &  \improv{88.54}{23.77}&  \improv{13.55}{14.7} & \improv{99.19}{8.36} &  \improv{63.53}{48.53} & \improv{3.76}{36.94} \\
         & LR  & \improv{98.37}{5.43} & \improv{92.86}{28.09} & \improv{8.95}{19.3} & \improv{99.47}{8.64} &  \improv{79.02}{64.02} & \improv{1.62}{39.08} \\
         \bottomrule   
        \end{tabular}}
        \caption{\textbf{Likelihood Ratio Gains Ablation.} We compare the OoD detection performance o different backbones and with different scoring functions on the Road Anomaly and FS LaF datasets. Gains/losses to the base ID scoring are highlighted in green/red, respectively.   
        } 
        \label{tab:llr_gains_ablation}
\end{table*}

% CLIP results: Road Anomaly:
%          idd_AUPR          │    0.5071387887001038     │
%│         idd_AUROC         │    0.9035064578056335     │
%│         idd_FPR95         │    0.3472384214401245     │
%│         llr_AUPR          │    0.8783584237098694     │
%│         llr_AUROC         │    0.9784472584724426     │
%│         llr_FPR95         │    0.1080690547823906     │
%│         ood_AUPR          │    0.8354361057281494     │
%│         ood_AUROC         │    0.9677878022193909     │
%│         ood_FPR95         │    0.17414529621601105
% Fishyscapes:
%         idd_AUPR          │    0.13281670212745667    │
%│         idd_AUROC         │    0.9197396039962769     │
%│         idd_FPR95         │    0.38695311546325684    │
%│         llr_AUPR          │    0.5583996772766113     │
%│         llr_AUROC         │    0.9894641637802124     │
%│         llr_FPR95         │    0.04122563451528549    │
%│         ood_AUPR          │    0.35701215267181396    │
%│         ood_AUROC         │    0.9855843186378479     │
%│         ood_FPR95         │    0.06053447350859642

\boldparagraph{Improvements from Likelihood Ratio}
 
We question whether the likelihood ratio is necessary for unknown segmentation. To investigate this, we push the performance of a generative model that requires no additional outlier training by using more powerful backbones. GMMSeg estimates class probability densities, allowing it to directly compute an anomaly score based on the likelihood of the maximum component without requiring outlier training. We use GMMSeg's density estimate (ID) as a baseline. We also consider the density estimate of the proxy OoD distribution alone as a scoring function (OoD). Lastly, we use the likelihood ratio scoring (LR), which integrates information from both distributions. 
Table \ref{tab:llr_gains_ablation} illustrates the performance improvements of the two scoring functions compared to the density estimates from the GMMSeg. The results consistently demonstrate that the likelihood ratio formulation provides better performance over inlier density estimates or the OoD scoring alone, highlighting the advantages of our approach. 

\begin{figure*}[tb]
  \centering
  \includegraphics[width=\linewidth]{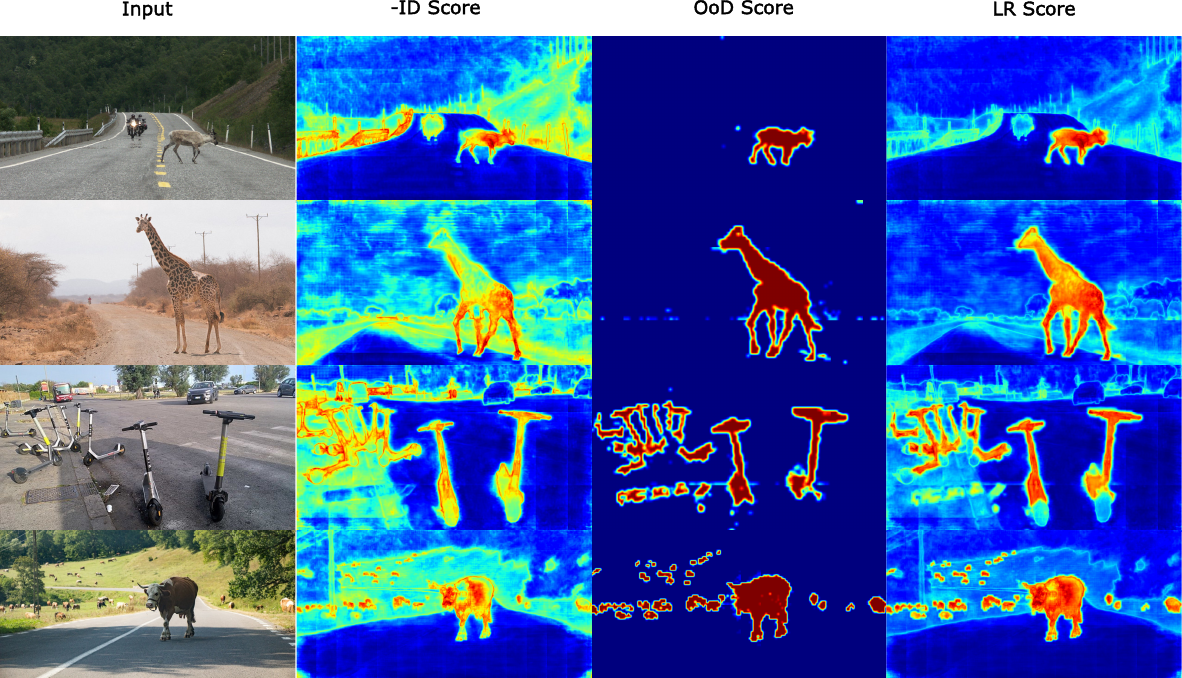}
  \caption{\textbf{Qualitative Results on SMIYC-AT (G-G).} The second column (-ID score) shows the outlier score from using the GMM without outlier supervision. The third column (OoD Score) shows the anomaly score from the fine-tuned OoD detection head. The fourth column (LR Score) shows our proposed likelihood formulation. The likelihood formulation combines information from both and predicts more accurate OoD score maps. 
  }
  \label{fig:qual}
\end{figure*}

We qualitatively compare the three scoring functions in \figref{fig:qual}.
The in-distribution (ID) score demonstrates lower precision due to its tendency to favor known classes. In contrast, the out-of-distribution (OoD) scoring detects outliers very confidently but at the cost of increasing false positives so as not to miss any outliers. The proposed likelihood ratio (LR) balances the two, leveraging the strengths of each to achieve the best results in terms of both inliers and outliers.

\begin{table*}[ht]
    \centering
    \adjustbox{max width=\textwidth}{%
    \begin{tabular}{l c c c c c c c c|| c c } 
    \toprule
    \multirow{2}{*}{Method}  & \multicolumn{2}{c}{SMIYC-AT}  & \multicolumn{2}{c}{SMIYC-OT} & \multicolumn{2}{c}{ RoadAnomaly}  & \multicolumn{2}{c||}{FS LaF} & \multicolumn{2}{c}{Average} \\
    \cmidrule(lr){2-3} \cmidrule(l){4-5} \cmidrule(lr){6-7} \cmidrule(l){8-9} \cmidrule(l){10-11}
           & $AP$ $\uparrow$ & $FPR$ $\downarrow$ & $AP$ $\uparrow$  & $FPR$ $\downarrow$ & $AP$ $\uparrow$ & $FPR$ $\downarrow$ & $AP$ $\uparrow$  & $FPR$ $\downarrow$ & $\overline{AP}$ $\uparrow$  & $\overline{FPR}$ $\downarrow$  \\
    \midrule
     cDNP~\cite{Galesso2023ICCVW}  &-&-&-&-& 79.78 & 18.18   & 69.80 &   7.50&-&- \\
    
    UGainS~\cite{Nekrasov2023GCPR} &-&-&-&-& \underline{88.98} & 10.42  & \underline{80.08} & 6.61 &-&- \\
     
     Maximized Entropy~\cite{Chan2021ICCV} & 85.47 & 15.00 & 85.07 & 0.75 &  - & - & 29.96 &  35.14 & - & -\\ 
     
     CSL~\cite{Zhang_AAAI_2024} & 80.08 & 7.16 & 87.10 & 0.67 & 61.38 & 43.80 & - & - & - & -\\

     Maskomaly~\cite{Ackermann2023BMVC}  & 93.35 & 6.87 & - & -  & 70.90 & 11.90 & - & -& - & -\\
     
     PEBAL~\cite{Tian2022ECCV} & 49.14 & 40.82 & 4.98 & 12.68 &  45.10 & 44.58 & 58.81 &  4.76 & 39.51 & 25.71\\
     
     Mask2Anomaly~\cite{Rai2023ICCV} & 88.72 & 14.63 & \underline{93.22} & \underline{0.20} & 79.70 & 13.45 & 46.04 & 4.36 & 76.92 & 8.16\\

     RbA~\cite{Nayal2023ICCV} & 90.90 & 11.60 & 91.80 & 0.50 & 85.42 & \underline{6.92} & 70.81 & 6.30 & 84.73 & 6.33\\ 

     EAM~\cite{Grcic23CVPRW}  & 93.75 & \textbf{4.09} & 92.87 &0.52 & 69.40 & 7.70 & \textbf{81.50} & \underline{4.20} & 84.38 & \underline{4.13} \\

     %UNO~\cite{delić2024ARKIV}  & 93.3  & 2.0 & 93.2 &0.2 & 88.5 & 7.4 & 81.8 & 1.3 & 89.2 & 2.7 \\
     
     \midrule
     UEM (G-G)& \underline{94.10} & 6.90 & 88.30 & 0.40 &
         
     93.75 & \textbf{6.32} & 71.34 & 6.04 & 86.87 & 4.92 \\

     UEM (G-D)& 92.50 & 11.30 & 92.0 & \underline{0.20} &
         
     \underline{92.86} & 8.95 & 79.02 & \textbf{1.62} & \textbf{89.10} & 5.52 \\

     UEM (D-D)& \textbf{95.60} & \underline{4.70} & \textbf{94.40} & \textbf{0.10} &
         
     90.94 & 8.03 & 72.83 & \underline{2.60} & \underline{88.44} & \textbf{3.86} \\

     \bottomrule
    \end{tabular}}
    \caption{\textbf{Quantitative Results on SMIYC-AT, SMIYC-OT, Road Anomaly, and FS LaF.} We compare our approach against existing OoD segmentation methods. Averaged across the four datasets, our approach sets a new state-of-the-art for both AP and FPR. The best result for each dataset is highlighted in \textbf{bold}, and the second best is \underline{underlined}. Our method UEM (X-Y) has the flexibility to use a discriminative~(D) or generative~(G) modeling for the inlier classification head (X) and unknown estimation module (Y). We report the results of three possible combinations.
 }
    \label{tab:pixel_metrics}
\end{table*} 
\boldparagraph{Comparison to State-of-the-Art}
\tabref{tab:pixel_metrics} shows our results compared to state-of-the-art methods on four datasets with the average performance over datasets in the last column.   
As each dataset has different characteristics, the existing methods behave differently across the datasets. The top-performing methods include recently proposed masked-based models RbA~\cite{Nayal2023ICCV}, EAM~\cite{Grcic2022ECCV}, and Mask2Anomaly~\cite{Rai2023ICCV}. While these methods achieve impressive performance in terms of accuracy, reasoning at the mask level hurts FPR, as considering a mask outlier introduces several false positives at once. Its negative effect on small objects can be seen by high FPR on SMIYC-OT and FS LaF. Our method achieves significantly lower FPR on these two datasets while being among the top-performing methods in terms of AP. Our method also achieves impressive accuracy levels on real-world images of SMIYC-OT and Road Anomaly, increasing AP by $1.85$ and $1.18$, respectively, without causing high FPR. 
Averaged across the four datasets in the last column, our method sets a new state-of-the-art in both metrics, outperforming the previous state-of-the-art by $3.71\%$ in AP and $0.27\%$ in FPR. 

Training data impacts the performance significantly. Both RbA and EAM are trained on Mapillary and Cityscapes datasets, whereas we train our inlier model only on Cityscapes. Additionally, EAM uses ADE20K~\cite{Zhou2017CVPR} for outlier supervision, which contains a broader range of classes than COCO. We only use COCO to ensure a fair comparison to other methods. We also note that outlier supervision used in most other methods negatively impacts the performance of the inlier segmentation network as reported in \cite{Tian2022ECCV, Nayal2023ICCV}.

\begin{table*}[ht]
    \centering
    \adjustbox{max width=\textwidth}{%
    \begin{tabular}{l l c c c c c } 
     \toprule
     \multirow{2}{*}{Backbone}  & \multicolumn{3}{c}{Road Anomaly} & \multicolumn{3}{c}{FS LaF}  \\
        \cmidrule(r){2-4} \cmidrule(l){5-7}
        & $AUROC$ $\uparrow$ & $AP$ $\uparrow$ & $FPR$ $\downarrow$ & $AUROC$ $\uparrow$ & $AP$ $\uparrow$ & $FPR$ $\downarrow$ \\
         \midrule
        \multirow{1}{*}{PEBAL~\cite{Tian2022ECCV}} & \improv{98.32}{10.69} & \improv{92.98}{47.88}  & \improv{7.13}{37.45} &  \decrease{98.95}{0.01} & \decrease{51.11}{7.7} & \improv{2.89}{1.87}   \\
        %\midrule
        \multirow{1}{*}{RbA~\cite{Nayal2023ICCV}} &
        \decrease{97.91}{0.08} & \improv{89.37}{3.95} & \decrease{10.96}{4.04} & \decrease{98.39}{0.23} &  \decrease{49.78}{21.03} & \improv{4.64}{1.66} \\
         %\midrule
        \multirow{1}{*}{UEM (Ours)} &
         98.08 & 92.86 & 8.95 & 98.75 &  79.02 & 1.62 \\ 
         \bottomrule   
        \end{tabular}}
        \caption{\textbf{Other Methods with DINOv2.} We compare the OoD performance of two other scoring functions using DINOv2 on Road Anomaly and FS LaF datasets. Gains/losses to the original method are highlighted in green/red, respectively. While DINOv2 improves the results of other methods, especially PEBAL on Road Anomaly, our method consistently achieves top results on both datasets, without affecting the inlier performance.}
        \label{tab:other_methods_with_dino}
\end{table*}

\boldquestion{Is DINOv2 All You Need} To assess the backbone's impact, we compare our approach to PEBAL~\cite{Tian2022ECCV} and RbA~\cite{Nayal2023ICCV}. We use their scoring functions to train our segmentation network with the DINOv2 backbone and adjust the outlier supervision process to their original implementations. 
As shown in \tabref{tab:other_methods_with_dino}, DINOv2 significantly improves PEBAL's performance on Road Anomaly across both metrics and results in a lower FPR on FS LaF. We can attribute these improvements to the more robust backbone.

For RbA, AP on Road Anomaly improves, but other metrics are better using the original model with Mask2Former. This is likely due to the implicit one \vs all behavior in mask classification models, which the RbA scoring function is specifically designed for. 
Finally, our outlier scoring function performs best overall without modifying any original network parameters, a critical constraint for real-world applications. Both other methods are reported to lose at least 1\% mIoU during fine-tuning. We found this effect exacerbated with DINO, which requires careful adjustments to mitigate inlier performance loss.
\boldparagraph{Discriminative \vs Generative Modeling of Estimator module} Our unknown estimation module models two distributions during fine-tuning. Each distribution can be modeled as a data density using generative GMMs or explicitly as a linear layer mapping function. 
In \tabref{tab:pixel_metrics}, we evaluated different possible combinations for each. We find both discriminative and generative classifiers to outperform the previous state-of-the-art methods, with the fully discriminative classifier for the OoD modeling being slightly better. We omit the discriminative inlier and generative outlier (D-G) combination as we find the GMM takes too long to converge due to the unbounded range of values coming from the MLP.

\boldparagraph{On the Number of Parameters in UEM} The original segmentation network consists of 101M parameters. Our UEM module introduces an additional 788K parameters, representing a less than 1\% increase in the overall model size. Despite this minimal parameter overhead, the UEM module significantly enhances OoD detection performance.

\section{Conclusion and Future Work}
\label{sec:conclusion}
In this work, we propose a novel strategy to utilize proxy outlier data for improved OoD detection without retraining the entire network. This allows us to build on the robust representation space of large foundational models, significantly enhancing the generalization capability of the proposed approach. We propose an unknown estimation module (UEM) that can be integrated into the existing segmentation networks to identify OoD objects effectively. We develop an OoD scoring function based on the likelihood ratio by combining UEM's outputs with inlier predictions. 
Our method sets a new state-of-the-art in outlier segmentation across multiple datasets, without causing any drops in the inlier performance.

For future work, we aim to investigate how the choice of proxy out-of-distribution (OoD) dataset influences the generalization performance of our method. In this study, we utilized the COCO dataset as the proxy OoD data for fair comparison with the other approaches.  
We plan to investigate the effect of mining more realistic outliers from real-world OoD datasets~\cite{Shoeb2024WACV} as future work.

%-------------------------------------------------------------------------

\backmatter

\bmhead{Acknowledgements}
This project is co-funded by KUIS AI, Royal Society Newton Fund Advanced Fellowship (NAF\textbackslash R\textbackslash 2202237), and the European Union (ERC, ENSURE, 101116486). Y.~Shoeb acknowledges funding from the German Federal Ministry for Economic Affairs and Climate Action within the project ``just better DATA''. We also thank Unvest R\&D Center for their support. Views and opinions expressed are however those of the author(s) only and do not necessarily reflect those of the European Union or the European Research Council. Neither the European Union nor the granting authority can be held responsible for them.

\newpage
\bibliographystyle{splncs04}
\bibliography{egbib}% common bib file

\end{document}